\documentclass[11pt]{article}

\usepackage[final]{acl}

\usepackage{times}
\usepackage{latexsym}
\usepackage{algorithm}
\usepackage{algorithmicx}
\usepackage{algpseudocode}

\usepackage{setspace}
\usepackage[T1]{fontenc}
\usepackage{amsmath}
\usepackage{amssymb}
\usepackage[utf8]{inputenc}
\usepackage{enumitem}
\usepackage{amsfonts}
\usepackage{multirow}
\usepackage{multicol}
\usepackage[most]{tcolorbox}
\usepackage{subcaption}  
\usepackage{soul}
\soulregister{\cite}7
\soulregister{\ref}7
\soulregister{\pageref}7
\usepackage{hyperref}
\usepackage{xcolor}
\definecolor{deepred}{RGB}{204,23,25}
\definecolor{deepgreen}{RGB}{0,100,0}
\definecolor{deepyellow}{RGB}{246,157,37}
\definecolor{lightgray}{gray}{0.95}
\definecolor{orange}{RGB}{255,140,0}
\definecolor{blue}{RGB}{70,130,180}
\usepackage[most]{tcolorbox}
\usepackage{tcolorbox}
\usepackage{booktabs}
\usepackage[utf8]{inputenc} 
\usepackage[T1]{fontenc}    
\usepackage{hyperref}       
\usepackage{url}            
\usepackage{booktabs}       
\usepackage{amsfonts}       
\usepackage{nicefrac}       
\usepackage{microtype}      
\usepackage{xcolor}         

\usepackage{bm}
\usepackage{hyperref}
\usepackage{url}
\usepackage{booktabs}
\usepackage{bbding}
\usepackage{wrapfig}
\usepackage{multirow}
\usepackage{makecell}
\usepackage{pifont}
\usepackage{enumitem}
\usepackage{colortbl}
\usepackage{amsmath}
\usepackage{amssymb}
\usepackage{tikz}
\usepackage{array}
\usepackage{tcolorbox}
\usepackage{subcaption}
\usepackage{tcolorbox}
\usepackage{minted}

\definecolor{TableRowColor}{HTML}{F3E6FA}

\usepackage[utf8]{inputenc}
\usepackage{dsfont}
\usepackage{bbm}

\usepackage{microtype}

\usepackage{inconsolata}

\usepackage{graphicx}

\title{Towards Robust Process Reward Modeling via Noise-aware Learning}

\author{
 \textbf{Bin Xie\textsuperscript{1,2}},
 \textbf{Bingbing Xu\textsuperscript{1}}$^{*}$,
 \textbf{Xueyun Tian\textsuperscript{1,2}},
 \textbf{Yilin Chen\textsuperscript{1,2}},
 \textbf{Huawei Shen\textsuperscript{1,2}},\\
\\
 \textsuperscript{1}State Key Laboratory of AI Safety, Institute of Computing Technology, CAS \\
 \textsuperscript{2}University of Chinese Academy of Sciences\\
 \texttt{\{xiebin23s@ict.ac.cn, xubingbing@ict.ac.cn\}}
}
\newcommand \footnoteONLYtext[1]
{
	\let \mybackup \thefootnote
	\let \thefootnote \relax
	\footnotetext{#1}
	\let \thefootnote \mybackup
	\let \mybackup \imareallyundefinedcommand
}
\begin{document}
\maketitle
\footnoteONLYtext{$^{*}$Corresponding author.}
\begin{abstract}
Process Reward Models (PRMs) have achieved strong results in complex reasoning, but are bottlenecked by costly process-level supervision. A widely used alternative, Monte Carlo Estimation (MCE), defines process rewards as the probability that a policy model reaches the correct final answer from a given reasoning step. However, step correctness is an intrinsic property of the reasoning trajectory, and should be invariant to policy choice. Our empirical findings show that MCE producing policy-dependent rewards that induce label noise, including false positives that reward incorrect steps and false negatives that penalize correct ones. To address above challenges, we propose a two-stage framework to mitigate noisy supervision. In the labeling stage, we introduce a reflection-aware label correction mechanism that uses a large language model (LLM) as a judge to detect reflection and self-correction behaviors related to the current reasoning step, thereby suppressing overestimated rewards. In the training stage, we further propose a \underline{\textbf{N}}oise-\underline{\textbf{A}}ware \underline{\textbf{I}}terative \underline{\textbf{T}}raining framework that enables the PRM to progressively refine noisy labels based on its own confidence. Extensive Experiments show that our method substantially improves step-level correctness discrimination, achieving up to a 27\% absolute gain in average F1 over PRMs trained with noisy supervision.

\end{abstract}
\section{Introduction}
Outcome reward models (ORMs) have driven remarkable success in advancing large language model (LLM) reasoning \cite{ahmadian2024back, shao2024deepseekmath, team2025kimi, guo2025deepseek}. However, they suffer from reward sparsity and struggle with credit assignment, leading to inefficient exploration and reward hacking in both reinforcement learning (RL) and test-time scaling \cite{sutton1998reinforcement, leike2018scalable, lightman2023let, cao2024enhancing, chan2024dense, qu2025survey}. 
Process reward models (PRMs) address these issues by providing step-by-step feedback, a technique that has been proven to be highly effective for complex reasoning tasks \cite{cobbe2021training, uesato2022solving, yu2023metamath, hao2024llm, wang2024math}.

PRMs, when trained on extensive labeled datasets, function as discriminative classifiers that evaluate the correctness of intermediate reasoning steps \cite{uesato2022solving, luo2024improve, zhang2025lessons}. However, a fundamental bottleneck for PRMs is the acquisition of high-quality labeled data. To this end, early efforts employed human annotators guided by detailed instructions and rigorous protocols to achieve reliable annotation quality \cite{lightman2023let}. Nonetheless, such human annotation is labor-intensive, costly, and difficult to scale, which has led to a growing interest in automated labeling approaches \cite{wang2024math, zheng2024processbench, sun2025efficient}.

\begin{figure}[t]
    \centering
    \includegraphics[width=0.98\linewidth]{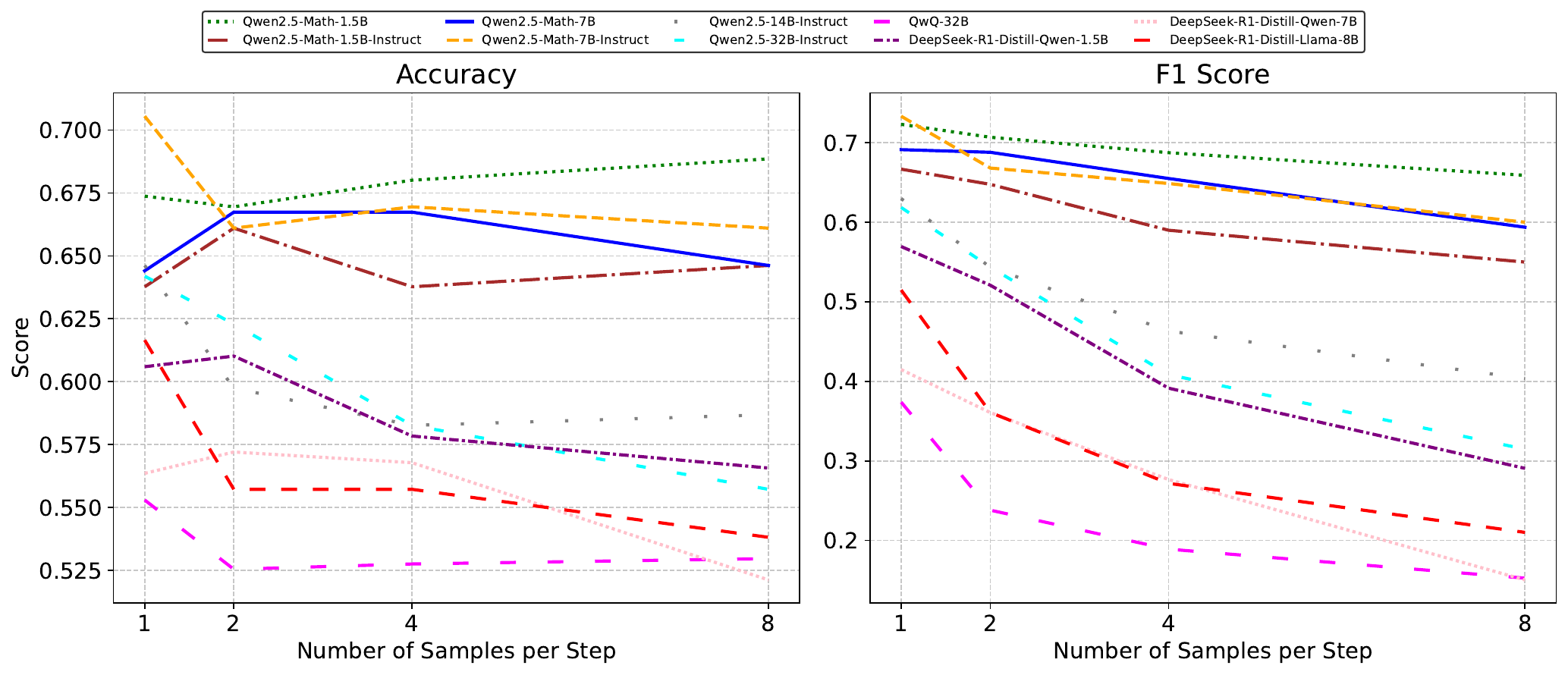}
    \caption{MCE method shows significant deviation compared to human-annotated ground truth, regardless of model capability and sampling budget.}
    \label{fig:correlation_plot}
\end{figure}

A representative method for automated labeling is Monte Carlo Estimation (MCE). It defines a step's reward as the probability of reaching a correct final answer, which is estimated by using a generative policy to sample numerous reasoning paths from the current step \cite{wang2024math, zhang2025lessons, luo2024improve, sun2025efficient}. However, step correctness is an intrinsic property of the reasoning trajectory, and should be invariant to policy choice. his highlights a fundamental conflict between MCE and process rewards, leading to significant label noise \cite{ding2025scan}. As shown in Figure~\ref{fig:correlation_plot}, MCE method correlate poorly with human annotations, regardless of the sampling budget or model capability.

To combat this, we first analyze the sources of MCE labeling errors, which we categorize as: 1) \textbf{False Positives (FP)}, where an incorrect step leads to a correct final answer due to the policy's powerful self-correction capabilities; and 2) \textbf{False Negatives (FN)}, where a correct step leads to an incorrect final answer due to subsequent failures in the reasoning path (as seen in Figure~\ref{fig:noise_illustration}). Motivated by these observations, we propose a two-stage approach to mitigate the impact of above noise on process reward modeling. First, during the data labeling phase, we design a reflection-aware label correction mechanism, which corrects overestimated rewards for steps involving self-correction, thereby addressing FPs. Second, during the model training phase, we propose a \underline{\textbf{N}}oise-\underline{\textbf{A}}ware \underline{\textbf{I}}terative \underline{\textbf{T}}raining framework. This framework allows the model effectively to combine the initial labels with its own confidence to correct the labels, thereby accurately judging the correctness of each reasoning step.

We conduct extensive experiments across diverse datasets and model scales. Compared to PRMs trained under noisy supervision, our model achieves up to a 27\% absolute improvement in average F1 for step-level correctness discrimination, while reaching comparable performance to strong critic models such as GPT-4o. Under test-time scaling with Best-of-$N$, our PRM further improves final answer accuracy by 3.2–5.6\% over baselines. Additional experiments show consistent gains as training data scales, indicating strong robustness and favorable scaling behavior. 

\section{Related Work}
\subsection{From Outcomes to Processes}
Process Reward Models (PRMs) provide fine-grained supervision over intermediate reasoning steps and have been shown to improve reasoning accuracy \cite{uesato2022solving, lightman2023let}. Most existing PRMs model step correctness via independent classification with cross-entropy loss \cite{cao2025process, han2025uncertainty, hu2025coarse, khalifa2025process, zhang2024entropy}, which ignores inter-step dependencies and is sensitive to noisy supervision. To better capture sequential structure, \citeauthor{li2024process} propose Process Q-value Models (PQM) that optimize relative Q-value rankings. Beyond discriminative formulations, Generative PRMs assign rewards conditioned on explicitly generated chain-of-thoughts, enabling stronger generalization and test-time scaling \cite{zeng2025optimizing, zhao2025genprm, zhang2025gm, zhang2025process}. In contrast, our work focuses on improving the robustness of discriminative PRMs under noisy process supervision.

\subsection{Data Annotation for Process Supervision}
Effective PRM training depends on high-quality process annotations, which are costly to obtain at scale. To reduce annotation overhead, prior work explores automated labeling via LLM-based annotators \cite{tan2025aurora, adak2025aura, wang2025test} or Monte Carlo Estimation (MCE), which estimates step correctness by sampling future reasoning trajectories \cite{wang2024math, sun2025efficient, luo2024improve}. However, MCE-based labels are highly sensitive to the sampling policy and often exhibit substantial noise. The most closely related work is \citealp{ding2025scan}, which reduces label noise via self-denoising across rollouts of a fixed policy, our approach directly addresses this limitation  by explicitly correcting reflection-induced overestimation and iteratively refining labels beyond a fixed policy.

\section{Preliminary}
\label{sec:preliminary}

\begin{figure}[!t]
    \centering
    \vspace{-10pt}
    \includegraphics[width=0.5\textwidth]{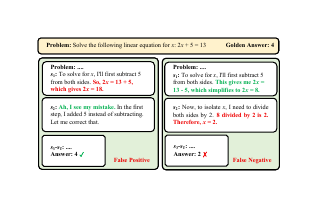}
    \vspace{-10pt}
    \caption{Illustration of MCE label noise. \textbf{(Left) False Positive:} A high reward is assigned to an incorrect step due to the policy's subsequent self-correction. \textbf{(Right) False Negative:} A low reward is assigned to a correct step due to a subsequent policy failure.}
    \label{fig:noise_illustration}
\end{figure}

We first review Monte Carlo Estimation (MCE) for process supervision, then formalize discriminative process reward modeling, and finally analyze how MCE induces policy-dependent label noise.

\begin{figure*}[t]
    \centering
    \includegraphics[width=\textwidth]{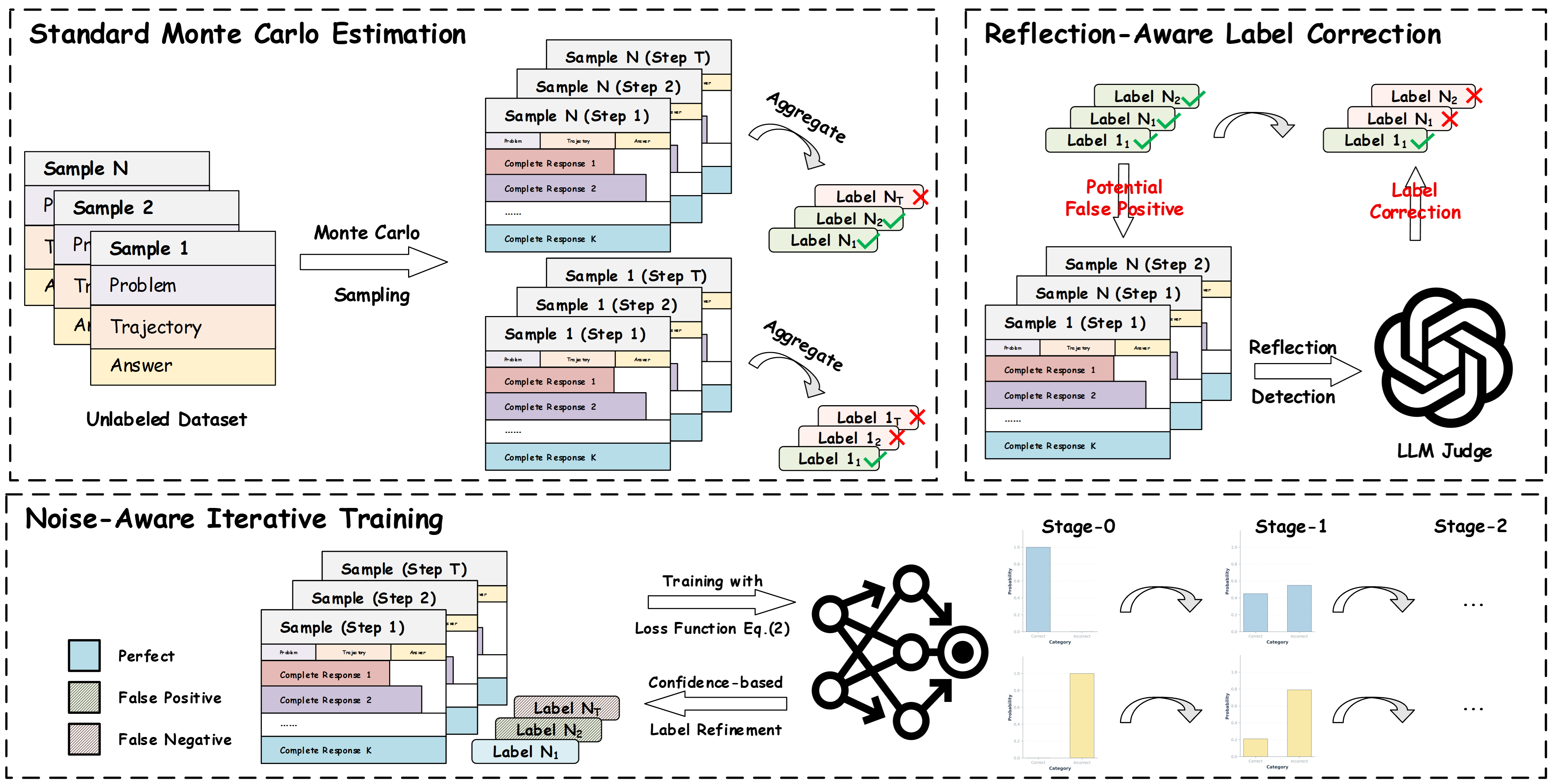}
    \caption{Overview of the proposed framework for robust process reward modeling. The framework consists of two stages. \textbf{(Top right)} In the data annotation stage, we propose a reflection-aware label correction mechanism, where an LLM judge detects reflection and self-correction behaviors within reasoning trajectories to identify and suppress overestimated rewards. \textbf{(Bottom)} In the training stage, we introduce a noise-aware iterative training framework that trains the process reward model with noise-robust objectives and progressively refines noisy labels based on model confidence across multiple stages.}
    \label{fig:overall_framework}
\end{figure*}

\subsection{Monte Carlo Estimation}
Given a generative policy $\pi$, MCE samples $K$ future continuations from step $s_t$, producing trajectories $\{\tau^{(k)}\}_{k=1}^K$, and compares their final answers $\mathrm{a}(\tau^{(k)})$ with the ground-truth answer $a^*$. Two common estimation strategies are used:
\begin{align}
    \hat{y}_t^{\text{SE}}(\pi) &= \frac{1}{K} \sum_{k=1}^{K} 
    \mathbbm{1}\bigl(\mathrm{a}(\tau^{(k)}) = a^*\bigr), \\
    \hat{y}_t^{\text{HE}}(\pi) &= \max_{k \in \{1,\ldots,K\}} \mathbbm{1}\bigl(\mathrm{a}(\tau^{(k)}) = a^*\bigr),
\end{align}
where $\mathbbm{1}(\cdot)$ denotes the indicator function. \textbf{Soft Estimation (SE)} assigns a probabilistic label, while \textbf{Hard Estimation (HE)} yields a binary label indicating whether at least one sampled continuation reaches the correct answer.

\subsection{Discriminative Process Reward Modeling}
Given a problem $P$ and a reasoning trajectory $\tau = (s_1, \ldots, s_T)$, a discriminative process reward model (PRM) $R_\theta$ assigns a correctness score to each step conditioned on its prefix:
\begin{equation}
    r_t = R_\theta(P, s_{\le t}), \quad r_t \in (0,1).
\end{equation}
Let $y_t \in [0,1]$ denote the step-level correctness label, which can be either binary (e.g., from HE) or continuous (e.g., from SE).
PRMs are typically trained using a cross-entropy objective:
\begin{equation}
    \mathcal{L}_{\text{CE}} = -\sum_{t=1}^{T} \left[ y_t \log r_t + (1-y_t)\log(1-r_t) \right].
\label{eq:ce}
\end{equation}
This formulation enables PRMs to act as discriminative classifiers that evaluate the validity of intermediate reasoning steps.

\subsection{Policy-Dependent Label Noise}
A fundamental limitation of MCE is that the estimated label $\hat{y}_t(\pi)$ depends on the \emph{future behavior of the sampling policy $\pi$}, rather than the intrinsic correctness of step $s_t$. This policy dependence introduces label noise, including: (i) \textbf{False Positives (FP)}, where incorrect steps receive high rewards due to downstream self-correction (As shown in Figure~\ref{fig:noise_illustration} Left), and (ii) \textbf{False Negatives (FN)}, where correct steps are penalized because later reasoning fails (As shown in Figure~\ref{fig:noise_illustration} Right). Mitigating such policy-induced noise is the central motivation of this work.

\section{Method}
\label{sec:method}
In this section, we propose a two-stage framework for training robust discriminative PRMs under noisy process supervision. Our method addresses policy-dependent noise introduced by MCE through (i) reflection-aware label correction during data annotation and (ii) noise-aware iterative training with selective label refinement. The overall workflow is illustrated in Figure~\ref{fig:overall_framework}.

\subsection{Reflection-Aware Label Correction}
As discussed in Section~\ref{sec:preliminary}, MCE assigns step-level labels by sampling $K$ trajectories from a generative policy conditioned on a given reasoning prefix. However, this procedure may produce false positives when an initially incorrect step is later corrected within a trajectory that still reaches the correct answer.

To mitigate this issue, we introduce a reflection-aware correction mechanism that operates at the \emph{trajectory level} prior to MCE aggregation. For each sampled trajectory $\tau^{(i)} = (s_1,\ldots,s_T)$, we apply a large language model to identify whether the correctness of the final answer is achieved by explicitly revising or negating an earlier step $s_t$. The prompt template is shown in Appendix~\ref{app:rd_prompt_template}. Formally, for a trajectory $\tau^{(i)}$ that reaches the correct final answer, we define a binary reflection indicator:
\begin{equation}
\textsc{Reflect}(\tau^{(i)}, t) \in \{0,1\},
\end{equation}
which evaluates whether any downstream step $s_k$ ($k>t$) explicitly corrects the content of $s_t$. If $\textsc{Reflect}(\tau^{(i)}, t)=1$, the trajectory is considered invalid evidence for the correctness of step $s_t$, and is re-labeled as unsuccessful with respect to $s_t$, even though it reaches the correct final answer.

After applying reflection detection to all sampled trajectories, MCE aggregation is performed on the corrected trajectory outcomes. This mechanism is designed as a high-precision filter that selectively removes spurious success trajectories caused by self-correction. As a result, it substantially reduces false positives introduced by MCE.

\subsection{Noise-Aware Iterative Training}
While reflection-aware correction substantially reduces false positives, residual label noise persists, particularly false negatives arising from limited policy capability. Directly correcting such cases via judgment-based supervision is inherently ambiguous, as downstream failures do not provide observable evidence that a preceding step is incorrect. To robustly train PRMs under this residual noise, we propose a noise-aware iterative training framework.

We first train an initial PRM $R_{\theta^{(0)}}$ on the filtered dataset with labels $\{\tilde{y}_t^{(0)}\}$ using standard cross-entropy loss. At iteration $k$, the model $R_{\theta^{(k)}}$ re-evaluates each step and outputs a continuous correctness score:
\begin{equation}
r_t^{(k)} = R_{\theta^{(k)}}(P, s_{\le t}), \quad r_t^{(k)} \in (0,1).
\end{equation}

For each step, we refine its supervision signal based on the discrepancy between the model prediction and the current label. Given a threshold $\delta$, the refined label $\tilde{y}_t^{(k+1)}$ is defined as:
\begin{equation}
\tilde{y}_t^{(k+1)} =
\begin{cases}
r_t^{(k)}, & \text{if } |r_t^{(k)} - \tilde{y}_t^{(k)}| > \delta, \\
\tilde{y}_t^{(k)}, & \text{otherwise}.
\end{cases}
\label{eq:iterative_refine}
\end{equation}

This selective refinement strategy preserves the majority of original labels while attenuating the influence of potentially corrupted ones. Under the empirically observed low-noise regime after reflection-based filtering, strong disagreement between the model prediction and the original label is more likely to indicate label noise than model error. The refinement and retraining procedure is repeated for multiple iterations to obtain the final PRM. The complete training procedure is summarized in Algorithm~\ref{alg:nait}.

\begin{algorithm}[t]
\caption{Noise-Aware Iterative Training}
\label{alg:nait}
\begin{algorithmic}[1]
\Require 
Filtered dataset $\mathcal{D}^{(0)}=\{(x_t, y_t^{(0)})\}_{t=1}^M$, 
threshold $\delta$, 
number of iterations $N$
\Ensure Trained process reward model $R_{\theta^{(N)}}$

\State Train initial PRM $R_{\theta^{(0)}}$ on $\mathcal{D}^{(0)}$

\For{$n = 0$ to $N-1$}
    \State Initialize refined dataset $\mathcal{D}^{(n+1)} \gets \emptyset$
    \For{each sample $(x_t, y_t^{(n)}) \in \mathcal{D}^{(n)}$}
        \State $r_t \gets R_{\theta^{(n)}}(x_t)$
        \If{$|r_t - y_t^{(n)}| > \delta$}
            \State $\tilde y_t \gets r_t$ 
        \Else
            \State $\tilde y_t \gets y_t^{(n)}$
        \EndIf
        \State Add $(x_t, \tilde y_t)$ to $\mathcal{D}^{(n+1)}$
    \EndFor
    \State Train PRM $R_{\theta^{(n+1)}}$ on $\mathcal{D}^{(n+1)}$
\EndFor

\State \Return $R_{\theta^{(N)}}$
\end{algorithmic}
\end{algorithm}

\section{Experiments}
\begin{table*}[!t]
    \centering
    \small
    \resizebox{\textwidth}{!}{\setlength{\tabcolsep}{1.4mm}{
    \begin{tabular}{l|ccc|ccc|ccc|ccc|c}
    \toprule
    \multirow{2}{*}{\bf Model} & \multicolumn{3}{c}{\bf GSM8K} & \multicolumn{3}{c}{\bf MATH} & \multicolumn{3}{c}{\bf \makecell{Olympiad Bench}} & \multicolumn{3}{c|}{\bf \makecell{Omni Math}} & \bf Avg. \\
    \cmidrule{2-13} & \textbf{error} & \textbf{correct} & \bf F1 & \textbf{error} & \textbf{correct} & \bf F1 & \textbf{error} & \textbf{correct} & \bf F1 & \textbf{error} & \textbf{correct} & \bf F1 & \bf F1 \\
    \midrule
    \multicolumn{14}{c}{\textbf{Criric Models (LLM-as-a-judge)}} \\
    \midrule
    GPT-4o-0806 & 70.0 & 91.2 & 79.2 & 54.4 & 76.6 & 63.6 & 45.8 & 58.4 & 51.4 & 45.2 & 65.6 & 53.5 & 61.9 \\
    Qwen2.5-72B-Instruct & 62.8 & 96.9 & 76.2 & 46.3 & 93.1 & 61.8 & 38.7 & 92.6 & 54.6 & 36.6 & 90.9 & 52.2 & 61.2 \\
    \midrule
    \multicolumn{14}{c}{\textbf{MCE-based PRMs}} \\
    \midrule
    Math-Shepherd-PRM-7B & 32.4 & 91.7 & 47.9 & 18.0 & 82.0 & 29.5 & 15.0 & 71.1 & 24.8 & 14.2 & 73.0 & 23.8 & 31.5 \\
    RLHFlow-PRM-Mistral-8B & 33.8 & 99.0 & 50.4 & 21.7 & 72.2 & 33.4 & 8.2 & 43.1 & 13.8 & 9.6 & 45.2 & 15.8 & 28.4 \\
    RLHFlow-PRM-Deepseek-8B & 24.2 & 98.4 & 38.8 & 21.4 & 80.0 & 33.8 & 10.1 & 51.0 & 16.9 & 10.9 & 51.9 & 16.9 & 26.6 \\
    $\bigstar$ Qwen2.5-Math-7B-\textsc{MC} & 33.2 & 92.7 & 48.7 & 21.5 & 77.8 & 32.5 & 12.6 & 60.1 & 20.6 & 12.9 & 54.6 & 19.9 & 30.4 \\
    \midrule
    \multicolumn{14}{c}{\textbf{Filtered-integrated PRMs}} \\
    \midrule
    Qwen2.5-Math-PRM-7B & 72.0 & 96.4 & \textbf{82.4} & 68.0 & 90.4 & \textbf{77.6} & 55.7 & 85.5 & \textbf{67.5} & 55.2 & 83.0 & \textbf{66.3} & \textbf{73.5} \\
    $\bigstar$ Qwen2.5-Math-7B-\textsc{MCRD} & 55.6 & 94.5 & 67.6 & 41.9 & 78.5 & 52.1 & 22.5 & 61.2 & 28.8 & 20.8 & 55.7 & 28.5 & 44.3 \\
    \midrule
    \multicolumn{14}{c}{\textbf{Other PRM Baselines}} \\
    \midrule
    Skywork-PRM-Qwen2.5-7B & 61.8 & 82.9 & 70.8 & 43.8 & 62.2 & 53.6 & 17.9 & 31.9 & 22.9 & 14.0 & 41.9 & 21.0 & 42.1 \\
    EurusPRM-Stage1 & 46.9 & 42.0 & 44.3 & 33.3 & 38.2 & 35.6 & 23.9 & 19.8 & 21.7 & 21.9 & 24.5 & 23.1 & 31.2 \\
    EurusPRM-Stage2 & 51.2 & 44.0 & 47.3 & 36.4 & 35.0 & 35.7 & 25.7 & 18.0 & 21.2 & 23.1 & 19.1 & 20.9 & 31.3 \\
    $\bigstar$ Qwen2.5-Math-7B-\textsc{NAIT} & 65.5 & 95.1 & \underline{77.9} & 55.2 & 79.8 & \underline{64.6} & 35.8 & 60.5 & \underline{44.1} & 32.9 & 61.9 & \underline{42.8} & \underline{57.4} \\
    \bottomrule
\end{tabular}
}}
\caption{Evaluation Results on ProcessBench. \textbf{Bold} represents the best result, \underline{Underline} represents the second best result. $\bigstar$ represents the models we trained. The results of \textsc{Qwen2.5-Math-1.5B-NAIT} are shown in Appendix~\ref{app:qwen1_5}.}
\label{tab:main_result_processbench}
\end{table*}
We conduct experiments to evaluate the effectiveness of our method at both the data and model levels.
Our evaluation focuses on the following research questions (RQs):

\begin{itemize}[leftmargin=*]
    \item \textbf{RQ1:} Can reflection-aware label correction improve the quality of the automated process-supervised annotation?
    \item \textbf{RQ2:} Can Noise-Aware Iterative Training enhance the PRMs performance?
\end{itemize}

\subsection{Experimental Setup}
We describe the dataset construction, compared baselines, and evaluation protocols. More experimental details are shown in Appendix~\ref{app:training_details}.

\begin{table}[!h]
\centering
\renewcommand{\arraystretch}{1}
\small
\setlength{\tabcolsep}{3pt}
\begin{tabular}{lcccccc}
    \toprule
    \textbf{Method} &
    \multicolumn{2}{c}{\textbf{MC-HE}} &
    \multicolumn{2}{c}{\textbf{7B-Inst.}} &
    \multicolumn{2}{c}{\textbf{14B-Inst.}} \\
    \cmidrule(lr){2-3} \cmidrule(lr){4-5} \cmidrule(lr){6-7}
    \textbf{Generator} & \textbf{Acc} & \textbf{F1} & \textbf{Acc} & \textbf{F1} & \textbf{Acc} & \textbf{F1} \\
    \midrule
    Math-1.5B.        & 0.655 & 0.698 & 0.669 & \textcolor{deepred}{\textbf{0.731}} & \textcolor{blue}{\textbf{0.669}} & 0.730 \\
    Math-1.5B-Inst.  & 0.665 & 0.650 & 0.691 & 0.717 & \textcolor{blue}{\textbf{0.706}} & \textcolor{deepred}{\textbf{0.724}} \\
    Math-7B.         & 0.646 & 0.674 & 0.672 & 0.726 & \textcolor{blue}{\textbf{0.684}} & \textcolor{deepred}{\textbf{0.732}} \\
    Math-7B-Inst.    & 0.665 & 0.671 & 0.693 & 0.729 & \textcolor{blue}{\textbf{0.703}} & \textcolor{deepred}{\textbf{0.730}} \\
    7B-Inst.         & 0.610 & 0.521 & 0.682 & 0.685 & \textcolor{blue}{\textbf{0.710}} & \textcolor{deepred}{\textbf{0.709}} \\
    14B-Inst.        & 0.610 & 0.551 & 0.678 & \textcolor{deepred}{\textbf{0.687}} & \textcolor{blue}{\textbf{0.684}} & 0.677 \\
    32B-Inst.        & 0.631 & 0.549 & 0.667 & \textcolor{deepred}{\textbf{0.680}} & \textcolor{blue}{\textbf{0.672}} & 0.634 \\
    \bottomrule
\end{tabular}
\caption{Label quality evaluation on the PRM800K test subset. We compare standard Monte Carlo Hard Estimation (\textbf{MC-HE}) against reflection-aware labeling with Qwen2.5-7B/14B-Instruct judges (\textbf{7B/14B-Inst.}), using various Qwen2.5-Math and Qwen2.5 series models as generators. \textcolor{blue}{\textbf{Blue}} and \textcolor{deepred}{\textbf{Red}} highlight the best Accuracy and F1 Score, respectively.}
\label{tab:reflection_label_quality}
\end{table}

\subsubsection{Training Datasets}

\paragraph{MC Dataset.}
We sample problems and ground-truth answers from the training splits of GSM8K, MATH-500, and PRM800K, and generate 128K reasoning trajectories (about 1M steps) using models from the \textsc{Qwen2.5} and \textsc{Qwen2.5-Math} Series. Step-level labels are produced via Monte Carlo Estimation with hard estimation and $K=8$, using \textsc{Qwen2.5-Math-7B-Instruct} as the sampling policy. The resulting dataset is referred to as \textbf{\emph{MC-7B}}.

\paragraph{MCRD Dataset.}
Starting from \textbf{\emph{MC-7B}}, we apply the proposed reflection-aware label correction to filter trajectories involving explicit self-correction before MCE aggregation, yielding the \textbf{\emph{MCRD}} dataset. We use \textsc{Qwen2.5-7B-Instruct} as the judge, producing the final dataset \textbf{\emph{MCRD-7B\_7B}}.

\begin{table*}[!t]
    \centering
    \small
    \resizebox{\textwidth}{!}{
    \begin{tabular}{l|cc|cccc|c}
    \toprule
    \bf Model & \bf \makecell{Training \\ Samples} & \bf \makecell{Aggregation \\ Method} & \bf GSM8K & \bf MATH & \bf \makecell{GAOKAO} & \bf \makecell{MINERVA \\ MATH} & \bf Avg. \\
    \midrule
    \multicolumn{2}{l}{Greedy} & & 94.4 & 78.0 & 69.3 & 49.2 & 72.7 \\
    \multicolumn{2}{l}{Majority Vote@8} & & 95.0 & 81.8 & 73.2 & 53.3 & 75.8 \\
    \multicolumn{2}{l}{\textbf{Pass@8 (Upper Bound)}} & & 97.0 & 87.0 & 80.0 & 63.2 & 81.8 \\
    \midrule
    Math-Shepherd-PRM-7B & 445K & Mean & 92.0 & 76.2 & 69.6 & 49.6 & 71.9 \\
    RLHFlow-PRM-Mistral-8B & 273K & Mean & 94.8 & 80.0 & 70.9 & 50.4 & 74.0 \\
    RLHFlow-PRM-DeepSeek-8B & 253K & Mean & 94.8 & 79.4 & 72.7 & 50.0 & 74.2 \\
    EurusPRM-Stage1 & 463K & Min & 93.5 & 79.4 & 71.2 & 52.6 & 74.2 \\
    EurusPRM-Stage2 & 693K & Sum & 93.9 & 80.6 & 72.2 & 51.5 & 74.6 \\
    Skywork-PRM-Qwen2.5-7B & - & Mean & 93.8 & 80.6 & \textbf{74.5} & 51.1 & 75.0 \\
    Qwen2.5-Math-PRM-7B & 1500K & Mean & \underline{95.4} & \underline{81.6} & \underline{73.5} & \textbf{54.0} & \textbf{76.1} \\
    $\bigstar$ Qwen2.5-Math-7B-\textsc{MC} & 128K & Mean & 91.8 & 75.9 & 69.1 & 49.3 & 71.5 \\
    $\bigstar$ Qwen2.5-Math-7B-\textsc{MCRD} & 128K & Mean & 95.0 & 80.2 & 72.9 & 50.8 & 74.7 \\
    $\bigstar$ Qwen2.5-Math-7B-\textsc{NAIT} & 128K & Mean & \textbf{95.6} & \textbf{82.1} & 73.2 & \underline{52.7} & \underline{75.9} \\
    \bottomrule
    \end{tabular}
    }
    \caption{Best-of-8 evaluation performance comparison on the Reasoning benchmarks with the policy model \textsc{Qwen2.5-14B-Instruct} in 3 scoring strategies. Mean, Min, and Sum represent the average/minimum/sum of each step in the reasoning trajectory as selection criteria, respectively. $\bigstar$ represents the PRMs we trained. The results of \textsc{Qwen2.5-Math-1.5B-NAIT} are shown in Appendix~\ref{app:qwen1_5}.}
    \label{tab:main_result_inference}
\end{table*}

\subsubsection{Baselines}
We categorize the baselines into three groups according to their labeling strategies and training paradigms.
\textbf{1) MCE-based PRMs.} These models are trained directly on datasets labeled using standard Monte Carlo Estimation, including Math-Shepherd \cite{wang2024math} and RLHFlow-PRM \cite{dong2024rlhf}. \textbf{2) Filtered-integrated PRMs.} These methods introduce additional label filtering or refinement mechanisms on top of MCE-labeled datasets, including Qwen2.5-Math-PRM \cite{zhang2025lessons}. \textbf{3) Other PRM Baselines.} We further include competitive PRMs obtained via alternative training pipelines or stronger supervision, namely Skywork-PRM \cite{he_2024_16998085} and Eurus-PRM \cite{cui2025process}.

\subsubsection{Evaluation Methods}
We evaluate our approach from three complementary perspectives.

\paragraph{Label Quality Evaluation.}
We assess the quality of automatically generated process labels on the PRM800K test set with human-annotated step-level correctness labels, reporting Accuracy and F1 score to compare standard MCE labeling with our reflection-aware correction.

\paragraph{Step-Level PRM Evaluation.}
We evaluate PRM performance on \textbf{ProcessBench}, reporting step-level Accuracy and F1 score to measure the ability to distinguish correct from incorrect reasoning steps across multiple mathematical reasoning datasets.

\paragraph{Test-Time Scaling Evaluation.}
We evaluate the practical utility of PRMs under test-time scaling using \emph{Best-of-$N$}, where PRM scores are used to select or guide reasoning trajectories, and report final answer accuracy on multiple mathematical reasoning tasks.

\subsection{Label Quality Evaluation (RQ1)}
\label{sec:rq1}

This section evaluates whether the proposed reflection-aware label correction can effectively reduce label noise introduced by MCE.

\paragraph{Main Results.}
We evaluate label quality on the PRM800K test set using human-annotated step-level correctness labels, reporting Accuracy and F1. As shown in Table~\ref{tab:reflection_label_quality}, reflection-aware correction consistently outperforms standard MC-HE across all generators. Compared to MC-HE, F1 improves by 3.3–16.4 points (e.g., 0.521 $\rightarrow$ 0.685 for \textsc{7B-Inst.}), with concurrent gains in Accuracy (e.g., 0.610 $\rightarrow$ 0.710 for \textsc{7B-Inst.}). Across most settings, \textsc{Qwen2.5-7B-Instruct} matches or slightly surpasses \textsc{Qwen2.5-14B-Instruct} in F1.

\paragraph{Analysis.}
These results support our hypothesis that downstream self-corrections are a reliable signal for identifying policy-induced false positives in MCE. By filtering such trajectories prior to aggregation, reflection-aware correction substantially improves precision, leading to large F1 gains while maintaining comparable recall. This high-precision behavior reduces label noise without sacrificing coverage of correct steps, yielding cleaner hard supervision. Overall, the improved label quality provides a stronger foundation for training robust PRMs in the subsequent noise-aware iterative training stage.

\subsection{PRMs Performance (RQ1 \& RQ2)}
This section evaluates the contribution of the proposed framework in Section~\ref{sec:method} to PRM training.

\begin{figure*}[!t]
    \centering
    \includegraphics[width=\linewidth]{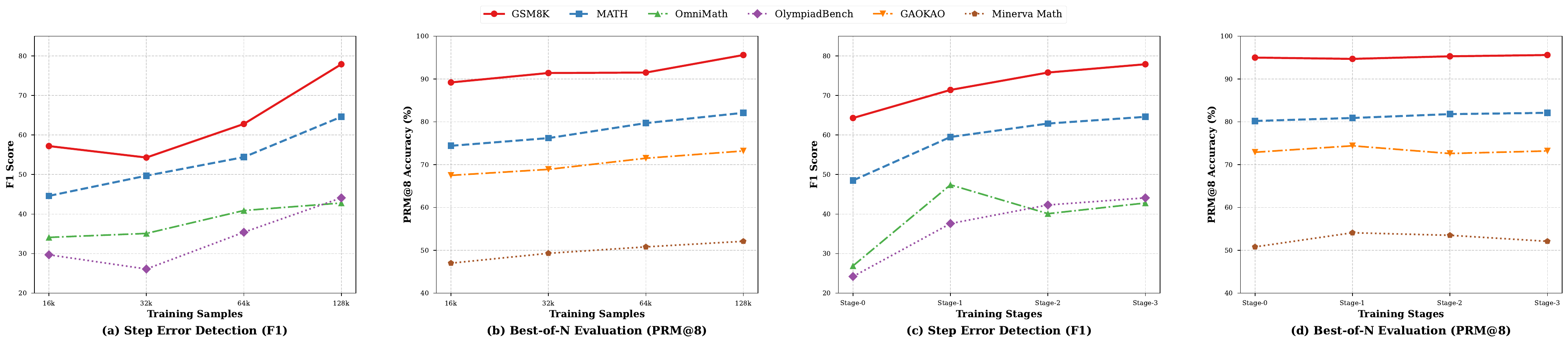}
    \caption{Effect of training data scale and iterative training on PRM performance. (a) The impact of data scale on PRM's F1 score on ProcessBench; (b) The impact of data scale on PRM's accuracy at PRM@8; (c) The impact of the number of training epochs on PRM's F1 score at ProcessBench; (d) The impact of the number of training epochs on PRM's accuracy at PRM@8.}
    \label{fig4v1}
\end{figure*}

\paragraph{Main Results.}
Table~\ref{tab:main_result_processbench} shows that reflection-aware correction (\textsc{MCRD}) consistently improves step-level F1 over MCE supervision across all benchmarks, confirming its effectiveness in suppressing policy-induced false positives. For example, compared to \textsc{MC}, \textsc{MCRD} improves average F1 from 30.4 to 44.3, with gains observed on all datasets. NAIT achieves further improvements, with an average F1 of 57.4, and approaches strong baselines such as GPT-4o (61.9), but using only a quarter or even less of the data. As shown in Table~\ref{tab:main_result_inference}, these improvements translate to downstream test-time scaling. Under Best-of-8 inference, \textsc{NAIT} improves average accuracy from 71.5 (\textsc{MC}) and 74.7 (\textsc{MCRD}) to 75.9, narrowing the gap to the Pass@8 upper bound.

\paragraph{Analysis.}
The results indicate that reflection-aware correction effectively sharpens step-level discrimination by reducing false positives introduced by MCE, yielding consistent F1 gains across datasets. \textsc{NAIT} further addresses residual noise, particularly false negatives, through iterative confidence-based label revision, leading to robust improvements without relying on large-scale supervision. Importantly, the alignment between step-level gains and Best-of-$N$ accuracy demonstrates that cleaner and more faithful process rewards directly improve trajectory ranking at inference time. Together, these components form a scalable framework for learning effective PRMs under noisy, automatically generated supervision.

\begin{figure}[!h]
    \centering
    \includegraphics[width=\linewidth]{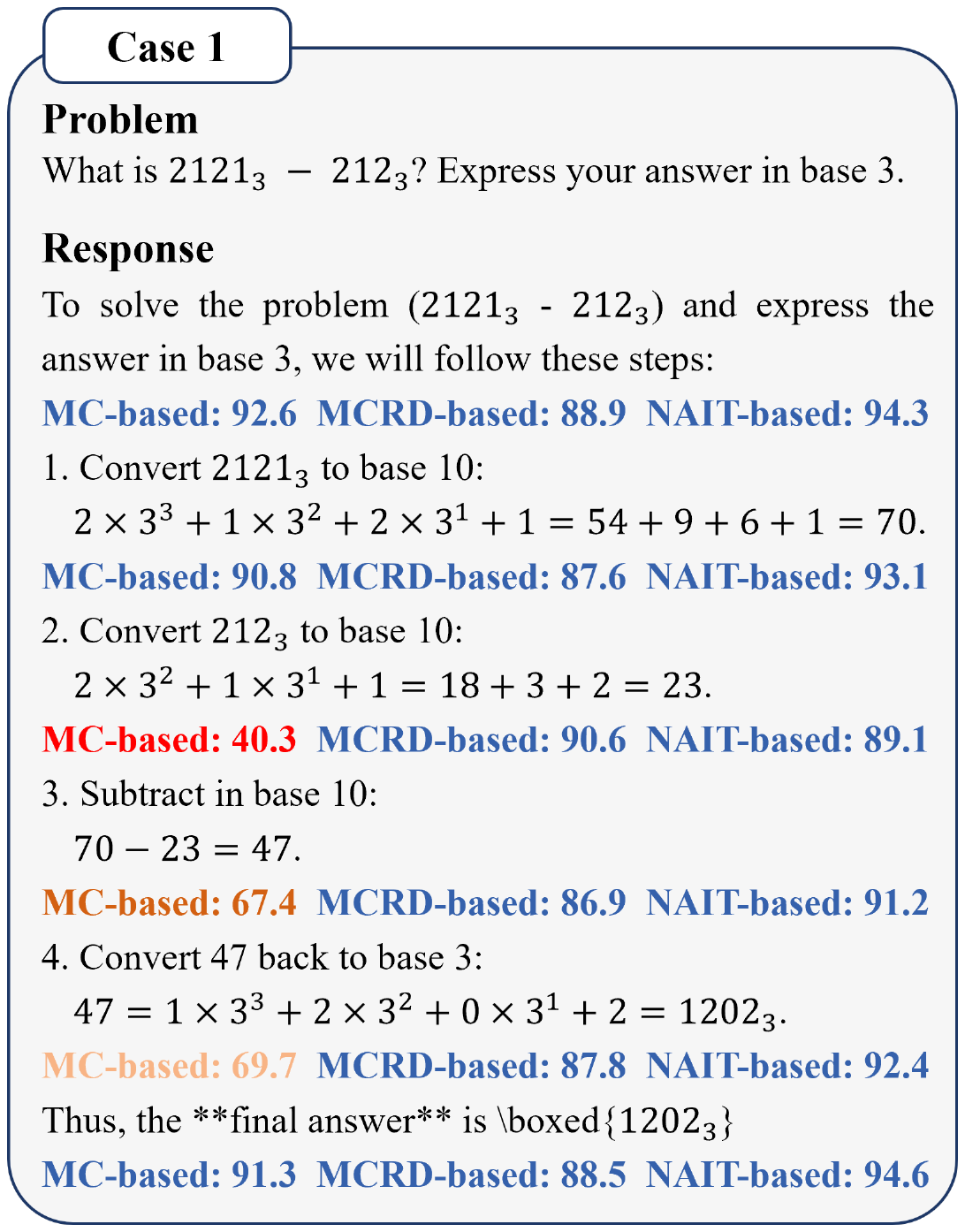}
    \caption{The MC-based PRM mislabels a correct step, whereas MCRD-based and NAIT-based PRMs assign high scores.}
    \label{fig5v1}
\end{figure}

\subsection{Further Analysis and Insights}
We further analyze the behavior of NAIT from two complementary perspectives: the effect of training data scale and the dynamics of iterative noise-aware optimization.

\paragraph{Effect of Training Data Scale.}
We study the impact of training data scale on PRM performance under the \textsc{NAIT} framework by training models on increasingly larger subsets of \emph{MCRD-7B\_7B} (16K–128K), with all other settings fixed. As shown in Figure~\hyperref[fig4v1]{4a}, step-level F1 improves consistently across all ProcessBench benchmarks as data scale increases, indicating stronger discrimination between correct and incorrect reasoning steps. For example, F1 on GSM8K rises from 57.2 to 77.9, while MATH and OmniMath improve from 44.6 to 64.6 and from 34.1 to 57.4, respectively, as training data scales up. These gains transfer directly to test-time reasoning. In Best-of-$N$ evaluation (Figure~\hyperref[fig4v1]{4b}, PRM@8 accuracy increases steadily with more training data, e.g., from 89.2 to 95.6 on GSM8K and from 74.4 to 82.1 on MATH, reflecting more reliable trajectory ranking and improved search guidance across tasks. These results demonstrate that the remaining gap to Qwen2.5-Math-PRM-7B is largely attributable to data scale rather than \textsc{NAIT} framework limitations.

\paragraph{Effect of Iterative Training in NAIT.}
We analyze the effect of iterative optimization in \textsc{NAIT} by evaluating PRMs across training stages (Stage~0–3) on the full \textsc{MCRD} dataset. Stage~0 uses the original hard labels, while Stage~1–3 apply Algorithm~\ref{alg:nait} with progressively smaller thresholds ($\tau=0.3, 0.2, 0.1$). As shown in Figure~\hyperref[fig4v1]{4c} and Figure~\hyperref[fig4v1]{4d}, performance improves monotonically across iterations on both ProcessBench and Best-of-$N$ evaluations. Step-level F1 increases substantially from Stage-0 to Stage-3 (e.g., GSM8K: 64.3$\rightarrow$77.9; MATH: 48.5$\rightarrow$64.6), indicating sharper step discrimination, while PRM@8 accuracy shows consistent but saturating gains at inference time (e.g., MATH: 80.2$\rightarrow$82.1). The diminishing improvements after later stages suggest that most recoverable label noise is corrected early, highlighting the effectiveness and stable convergence of iterative noise-aware refinement.

\subsection{Case Study}
We analyze two representative reasoning trajectories to qualitatively compare step-level rewards assigned by different PRMs, illustrating how our framework mitigates policy-dependent noise in MCE.

\paragraph{Case 1.} As shown in Figure~\ref{fig5v1} highlights a typical false negative case. Although the step ($212_3 \!\to\! 23$) is mathematically and logically correct, the MC-based PRM erroneously assigns a disproportionately low score, starkly reflecting its inherent sensitivity to the stochastic limitations of the sampling policy rather than the step's intrinsic correctness. In contrast, the NAIT-based PRM assigns a confidently and consistently high reward to this valid step, robustly correcting the misleading, noisy supervision. This compellingly demonstrates that iterative, noise-aware training significantly enables the PRM to autonomously override spurious failure signals derived from imperfect rollouts.

\begin{figure}[!h]
    \centering
    \includegraphics[width=\linewidth]{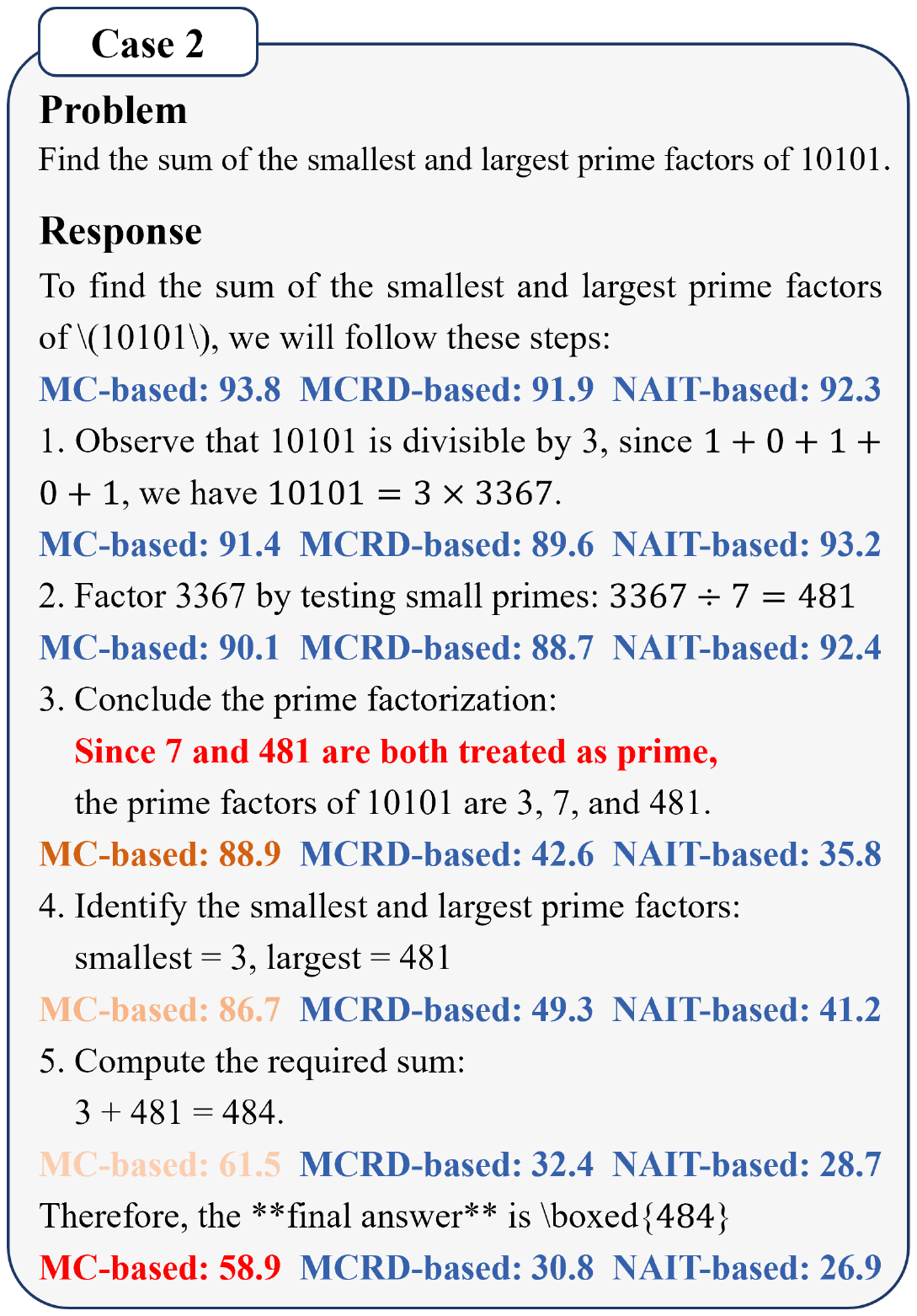}
    \caption{The MC-based PRM assigns high scores to an incorrect step and its consequences, while MCRD-based and NAIT-based PRMs suppress this effect.}
    \label{fig6v1}
\end{figure}

\paragraph{Case 2.} As shown in Figure~\ref{fig6v1} highlights a typical false positive case. The MC-based PRM assigns a high score to an erroneous claim (“$481$ is prime”). Both MCRD-based and NAIT-based PRMs sharply penalize this step, indicating effective suppression of false positives via reflection-aware correction. Moreover, while the MC-based PRM continues to assign high scores to subsequent steps conditioned on the incorrect premise, the NAIT-based PRM propagates the penalty forward, consistently assigning low rewards to downstream steps.

\section{Conclusion}
In this paper, we study the problem of noisy supervision in process reward modeling and analytically show that Monte Carlo Estimation inevitably introduces policy-dependent errors that fundamentally hinder PRM training. We propose a sophisticated, two-stage solution that seamlessly combines reflection-aware label correction during data construction with Noise-Aware Iterative Training (NAIT) during optimization, effectively enabling PRMs to precisely correct both deceptive false positives and overlooked false negatives without costly additional supervision. Extensive and rigorous experiments demonstrate that our approach consistently and significantly improves step-level discrimination and robust test-time reasoning performance across diverse datasets and varying model scales, while achieving remarkable data efficiency.
\section*{Limitations}
The process reward model trained in this study was fine-tuned based on \textsc{Qwen2.5-Math-1.5B}\footnote{\href{https://huggingface.co/Qwen/Qwen2.5-Math-1.5B}{Qwen/Qwen2.5-Math-1.5B}} and \textsc{Qwen2.5-Math-7B}\footnote{\href{https://huggingface.co/Qwen/Qwen2.5-Math-7B}{Qwen/Qwen2.5-Math-7B}}. The results may not be applicable to larger-scale models. Due to computational limitations, only test-time scaling was used to test the performance of PRM on downstream tasks, and it was not extended to reinforcement learning training.



\bibliography{custom}
\newpage
\appendix

\section{Training Details}
\label{app:training_details}
\paragraph{Software and hardware.} We conduct our experiments on a server with NVIDIA 4xL20 GPUs (48GB VRAM) and NVIDIA 4xA800 GPUs (80GB VRAM). We use Ubuntu 22.04.2 LTS as the operating system, with NVIDIA CUDA Toolkit version 12.4. All experiments are implemented in Python 3.10.15 using the PyTorch\footnote{\href{https://pytorch.org/}{PyTorch}} v2.6.0, OpenRLHF\footnote{\href{https://github.com/OpenRLHF/OpenRLHF}{OpenRLHF/OpenRLHF}} v0.9.1 and vLLM\footnote{\href{https://github.com/vllm-project/vllm}{vllm-project/vllm}} v0.8.4 packages.

\paragraph{Dataset Construction.} We sample problems and ground-truth answers from the training splits of GSM8K, MATH-500, and PRM800K, and generate 128K reasoning trajectories (about 1M steps) using models from the \textsc{Qwen2.5} and \textsc{Qwen2.5-Math} Series. Step-level labels are produced via Monte Carlo Estimation with hard estimation and $K=8$, using \textsc{Qwen2.5-Math-7B-Instruct} as the sampling policy. The resulting dataset is referred to as \textbf{\emph{MC-7B}}. Starting from \textbf{\emph{MC-7B}}, we apply the proposed reflection-aware label correction to filter trajectories involving explicit self-correction before MCE aggregation, yielding the \textbf{\emph{MCRD}} dataset. We use \textsc{Qwen2.5-7B-Instruct} as the judge, producing the final dataset \textbf{\emph{MCRD-7B\_7B}}.



\paragraph{Implementation Details.} During the training of PRM, we continued the language modeling approach. Specifically, we introduced three additional special tokens into the model's vocabulary: [PRM], [POS], and [NEG]. During training, we added [PRM] as a placeholder token after each reasoning step. When calculating the loss, the model masked the original text tokens and only output the logits corresponding to the last token of each step, namely [POS] and [NEG]. Then, we used Equation~\ref{eq:ce} to calculate the cross-entropy loss. The advantage of this method is that it can better adapt to the data distribution learned by the pre-trained model from a large-scale corpus, avoiding severe distribution shifts during training that could lead to a loss of basic semantic understanding capabilities.

\paragraph{Hyperparameters Specification.} We further fine-tune based on Qwen2.5-Math-1.5B and Qwen2.5-Math-7B to obtain our process reward model. The hyperparameters for Qwen2.5-Math-1.5B-NAIT and \textsc{Qwen2.5-Math-7B-NAIT} are shown in Table~\ref{tab:qwen1_5_nait} and the Table~\ref{tab:qwen7_nait}.

\begin{table}[h]
\centering
\small
\renewcommand{\arraystretch}{1}
\setlength{\tabcolsep}{2pt}
\begin{tabular}{lll}
\toprule
\textbf{Model} & \textbf{Parameter} & \textbf{Value} \\
\midrule
\multirow{7}{*}{{Qwen2.5-Math-1.5B-NAIT}} & LR & 5e-6 \\
& Number of Epochs & 1 (per stage)\\
& Gradient Acc. Steps & 128 \\
& DeepSpeed Zero Stage & 2 \\
& Batch Size & 1024 \\
& Optimizer & AdamW \\
& LR Scheduler & Cosine \\
& Max Length & 4096 \\
\bottomrule
\end{tabular}
\caption{Training Hyperparameters for \textsc{Qwen2.5-Math-1.5B-NAIT}}
\label{tab:qwen1_5_nait}
\end{table}

\begin{table}[h]
\centering
\small
\renewcommand{\arraystretch}{1}
\setlength{\tabcolsep}{2pt}
\begin{tabular}{lll}
\toprule
\textbf{Model} & \textbf{Parameter} & \textbf{Value} \\
\midrule
\multirow{7}{*}{Qwen2.5-Math-7B-NAIT} & LR & 3e-6 \\
& Number of Epochs & 1 (per stage)\\
& Gradient Acc. Steps & 128 \\
& DeepSpeed Zero Stage & 3 \\
& Batch Size & 1024 \\
& Optimizer & AdamW \\
& LR Scheduler & Cosine \\
& Max Length & 4096 \\
\bottomrule
\end{tabular}
\caption{Training Hyperparameters for \textsc{Qwen2.5-Math-7B-NAIT}}
\label{tab:qwen7_nait}
\end{table}

\section{Reflection-Aware Label Correction Prompt Template}
\label{app:rd_prompt_template}
The reflection-Aware label correction prompt template is shown in Figure~\ref{fig:rd_prompt}.

\begin{figure*}[h]
\centering
\begin{tcolorbox}[colback=lightgray, colframe=black, title=Reflection-Aware Label Correction Prompt Template]
\begin{spacing}{1.05}
\textbf{[SYSTEM PROMPT]} \newline
You are an expert evaluator of reasoning processes. I will provide a math problem, the history of reasoning steps, the current step of the solution and the reasoning trajectory starting from the current step) that leads to a final answer. They will be formatted as follows.

\textbf{[USER PROMPT]}

[Math Problem]

<math\_problem>\newline
...(math problem)...\newline
</math\_problem>

[History Steps]

<history\_step\_1>\newline
...(history step 1 of the solution)...\newline
</history\_step\_1>

[Current Step]

<current\_step>\newline
...(current step)...\newline
</current\_step>

[Reasoning Trajectory]

<reasoning\_trajectory>\newline
...(trajectory)...\newline
</reasoning\_trajectory>

Your task is to determine whether [Current Step] is actually correct, even if the subsequent reasoning process starting from it eventually reaches the correct final answer. You need to provide the analysis and the conclusion in the following format:

<analysis>
...(analysis of the current step)...
</analysis>

<conclusion>
Correct/Incorrect
</conclusion>

* Analyze [Current Step] in relation to [Math Problem], [History Steps], and the subsequent [Reasoning Trajectory].

* Your primary goal is to determine if [Current Step] is 'Correct' or 'Incorrect'.

* Examine [Reasoning Trajectory] for any reflection, self-correction, or revision.

* Specifically, identify if such corrections in [Reasoning Trajectory] address an error, flaw, significant omission, logical inconsistency (with [History Steps] or [Math Problem]), or misleading direction that originated in or was caused by [Current Step] itself.

* Label [Current Step] as 'Incorrect' if [Reasoning Trajectory] corrected a flaw that was present in [Current Step].

* Label [Current Step] as 'Correct' if it is logically sound and mathematically valid (given [Math Problem] and [History Steps]), AND [Reasoning Trajectory] builds upon it without needing to correct a fundamental flaw within [Current Step] itself.

* Note: Minor self-corrections within the [Reasoning Trajectory] that are unrelated to fixing a flaw in [Current Step] do not make [Current Step] incorrect.

* Respond with your analysis and conclusion directly.

\end{spacing}
\end{tcolorbox}
\caption{Reflection-Aware Label Correction Prompt Template}
\label{fig:rd_prompt}
\end{figure*}

\section{Qwen2.5-Math-7B Results}
\label{bon_evaluation}
We conducted experiments on policy models with different parameter sizes in Best-of-N Evaluation and compared our trained \textsc{Qwen2.5-Math-7B} model with other baselines. The results are shown in Tables ~\ref{tab:main_result_inference_3b}, Table ~\ref{tab:main_result_inference_7b}, and Table ~\ref{tab:main_result_inference_32b}.

\begin{table*}[!h]
    \centering
    \small
    \resizebox{\textwidth}{!}{
    \begin{tabular}{l|cc|cccc|c}
    \toprule
    \bf Model & \bf \makecell{Training \\ Samples} & \bf \makecell{Aggregation \\ Method} & \bf GSM8K & \bf MATH & \bf \makecell{GAOKAO} & \bf \makecell{MINERVA \\ MATH} & \bf Avg. \\
    \midrule
    \multicolumn{2}{l}{Greedy} & & 85.2 & 66.8 & 60.0 & 27.2 & 59.8 \\
    \multicolumn{2}{l}{Majority Vote@8} & & 86.0 & 74.0 & 66.5 & 38.6 & 66.3 \\
    \multicolumn{2}{l}{\textbf{Pass@8 (Upper Bound)}} & & 93.4 & 75.4 & 69.6 & 46.0 & 71.1 \\
    \midrule
    Math-Shepherd-PRM-7B & 445K & Mean & 86.6 & 63.6 & 58.7 & 33.8 & 60.7 \\
    RLHFlow-PRM-Mistral-8B & 273K & Mean & 89.4 & 65.2 & 59.5 & 36.4 & 62.6 \\
    RLHFlow-PRM-DeepSeek-8B & 253K & Mean & 89.2 & 66.2 & 59.2 & 34.9 & 62.4 \\
    EurusPRM-Stage1 & 463K & Min-Max & 88.0 & 67.6 & 60.2 & 36.4 & 63.1 \\
    EurusPRM-Stage2 & 693K & Sum & 88.6 & 67.4 & 60.8 & 35.6 & 63.1 \\
    Skywork-PRM-Qwen2.5-7B & - & Mean & 89.5 & \textbf{70.6} & \underline{63.1} & \textbf{38.2} & \underline{65.4} \\
    Qwen2.5-Math-PRM-7B & 1500K & Mean & \underline{90.8} & 70.4 & \textbf{63.6} & 37.5 & \textbf{65.6} \\
    $\bigstar$ Qwen2.5-Math-7B-\textsc{MC} & 128K & Mean & 86.4 & 63.3 & 58.4 & 33.5 & 60.4 \\
    $\bigstar$ Qwen2.5-Math-7B-\textsc{MCRD} & 128K & Mean & 89.8 & 66.5 & 59.9 & 36.7 & 63.2 \\
    $\bigstar$ Qwen2.5-Math-7B-\textsc{NAIT} & 128K & Mean & \textbf{91.2} & \underline{70.5} & 62.8 & \underline{37.1} & 65.4 \\
    \bottomrule
    \end{tabular}
    }
    \caption{Best-of-8 evaluation results of the policy model Qwen2.5-3B-Instruct. The best results among PRMs are \textbf{bolded} and the second best are \underline{underlined}.}
    \label{tab:main_result_inference_3b}
\end{table*}

\begin{table*}[!h]
    \centering
    \small
    \resizebox{\textwidth}{!}{
    \begin{tabular}{l|cc|cccc|c}
    \toprule
    \bf Model & \bf \makecell{Training \\ Samples} & \bf \makecell{Aggregation \\ Method} & \bf GSM8K & \bf MATH & \bf \makecell{GAOKAO} & \bf \makecell{MINERVA \\ MATH} & \bf Avg. \\
    \midrule
    \multicolumn{2}{l}{Greedy} & & 91.2 & 74.0 & 67.0 & 43.0 & 68.8 \\
    \multicolumn{2}{l}{Majority Vote@8} & & 92.9 & 78.6 & 70.1 & 46.7 & 72.1 \\
    \multicolumn{2}{l}{\textbf{Pass@8 (Upper Bound)}} & & 95.0 & 84.2 & 77.4 & 57.7 & 78.5 \\
    \midrule
    Math-Shepherd-PRM-7B & 445K & Mean & 93.2 & 76.6 & 67.0 & 44.8 & 70.4 \\
    RLHFlow-PRM-Mistral-8B & 273K & Mean & 93.0 & 76.6 & 67.0 & 42.3 & 69.7 \\
    RLHFlow-PRM-DeepSeek-8B & 253K & Mean & 93.7 & 76.8 & 68.3 & 44.8 & 70.9 \\
    EurusPRM-Stage1 & 463K & Min-Max & 93.8 & 79.6 & 70.3 & 44.8 & 72.1 \\
    EurusPRM-Stage2 & 693K & Sum & 94.0 & 79.2 & 71.4 & 45.6 & 72.6 \\
    Skywork-PRM-Qwen2.5-7B & - & Mean & 94.3 & \textbf{80.6} & \textbf{72.2} & \textbf{48.5} & \textbf{73.9} \\
    Qwen2.5-Math-PRM-7B & 1500K & Mean & \underline{94.5} & 79.8 & 71.4 & \underline{46.7} & \underline{73.1} \\
    $\bigstar$ Qwen2.5-Math-7B-\textsc{MC} & 128K & Mean & 93.1 & 76.2 & 66.8 & 44.2 & 70.1 \\
    $\bigstar$ Qwen2.5-Math-7B-\textsc{MCRD} & 128K & Mean & 93.8 & 77.5 & 68.6 & 45.1 & 71.2 \\
    $\bigstar$ Qwen2.5-Math-7B-\textsc{NAIT} & 128K & Mean & \textbf{94.8} & \underline{80.1} & \underline{71.3} & 45.8 & 73.0 \\
    \bottomrule
    \end{tabular}
    }
    \caption{Best-of-8 evaluation results of the Qwen2.5-7B-Instruct. The best results among PRMs are \textbf{bolded} and the second best are \underline{underlined}.}
    \label{tab:main_result_inference_7b}
\end{table*}

\begin{table*}[!h]
    \centering
    \small
    \resizebox{\textwidth}{!}{
    \begin{tabular}{l|cc|cccc|c}
    \toprule
    \bf Model & \bf \makecell{Training \\ Samples} & \bf \makecell{Aggregation \\ Method} & \bf GSM8K & \bf MATH & \bf \makecell{GAOKAO} & \bf \makecell{MINERVA \\ MATH} & \bf Avg. \\
    \midrule
    \multicolumn{2}{l}{Greedy} & & 95.7 & 82.8 & 77.1 & 55.9 & 77.9 \\
    \multicolumn{2}{l}{Majority Vote@8} & & 94.1 & 79.2 & 73.8 & 53.3 & 75.1 \\
    \multicolumn{2}{l}{\textbf{Pass@8 (Upper Bound)}} & & 96.6 & 85.6 & 79.2 & 62.1 & 80.9 \\
    \midrule
    Math-Shepherd-PRM-7B & 445K & Mean & 91.2 & 72.4 & 67.3 & 48.9 & 70.0 \\
    RLHFlow-PRM-Mistral-8B & 273K & Mean & 95.3 & 79.2 & 71.4 & 52.6 & 74.6 \\
    RLHFlow-PRM-DeepSeek-8B & 253K & Mean & \underline{95.4} & 78.0 & 72.5 & 52.9 & 74.7 \\
    EurusPRM-Stage1 & 463K & Min-Max & 94.1 & 78.6 & 70.4 & 49.6 & 73.2 \\
    EurusPRM-Stage2 & 693K & Sum & 92.6 & 78.6 & 69.1 & 50.7 & 72.8 \\
    Skywork-PRM-Qwen2.5-7B & - & Mean & 94.5 & 81.4 & 73.5 & 50.4 & 75.0 \\
    Qwen2.5-Math-PRM-7B & 1500K & Mean & 95.2 & \textbf{82.6} & \textbf{74.8} & \textbf{54.4} & \textbf{76.8} \\
    $\bigstar$ Qwen2.5-Math-7B-\textsc{MC} & 128K & Mean & 91.0 & 72.1 & 67.1 & 48.5 & 69.7 \\
    $\bigstar$ Qwen2.5-Math-7B-\textsc{MCRD} & 128K & Mean & 95.3 & 78.8 & 72.9 & 53.2 & 75.1 \\
    $\bigstar$ Qwen2.5-Math-7B-\textsc{NAIT} & 128K & Mean & \textbf{95.8} & \underline{82.5} & \underline{74.1} & \underline{53.6} & \underline{76.5} \\
    \bottomrule
    \end{tabular}
    }
    \caption{Best-of-8 evaluation results of the policy model Qwen2.5-32B-Instruct. The best results among PRMs are \textbf{bolded} and the second best are \underline{underlined}.}
    \label{tab:main_result_inference_32b}
\end{table*}

\section{Qwen2.5-Math-1.5B Results}
\label{app:qwen1_5}
Similarly, we also fine-tuned the \textsc{Qwen2.5-Math-1.5B} model on the MC and MCRD datasets using the proposed NAIT framework. The evaluation results are shown in Tables ~\ref{tab:qwen1_5_processbench}, Table ~\ref{tab:qwen1.5_3b}, Table ~\ref{tab:qwen1.5_7b}, Table ~\ref{tab:qwen1.5_14b}, and Table ~\ref{tab:qwen1.5_32b}.

\begin{table*}[!h]
    \centering
    \small
    \resizebox{\textwidth}{!}{\setlength{\tabcolsep}{1.4mm}{
    \begin{tabular}{l|ccc|ccc|ccc|ccc|c}
    \toprule
    \multirow{2}{*}{\bf Model} & \multicolumn{3}{c}{\bf GSM8K} & \multicolumn{3}{c}{\bf MATH} & \multicolumn{3}{c}{\bf \makecell{Olympiad Bench}} & \multicolumn{3}{c|}{\bf \makecell{Omni Math}} & \bf Avg. \\
    \cmidrule{2-13} & \textbf{error} & \textbf{correct} & \bf F1 & \textbf{error} & \textbf{correct} & \bf F1 & \textbf{error} & \textbf{correct} & \bf F1 & \textbf{error} & \textbf{correct} & \bf F1 & \bf F1 \\
    \midrule
    Qwen2.5-Math-1.5B-\textsc{MC} & 29.7 & 89.3 & 42.5 & 17.2 & 76.9 & 27.3 & 11.4 & 65.9 & 19.5 & 8.1 & 49.3 & 14.9 & 26.0 \\
    Qwen2.5-Math-1.5B-\textsc{MCRD} & 39.5 & 76.2 & 43.5 & 31.7 & 70.4 & 40.2 & 17.1 & 57.6 & 22.8 & 16.9 & 47.6 & 23.1 & 32.4 \\
    Qwen2.5-Math-1.5B-\textsc{NAIT} & 51.4 & 77.3 & 60.1 & 36.8 & 68.2 & 46.1 & 16.5 & 54.8 & 21.4 & 19.5 & 50.9 & 28.4 & 39.0 \\
    \bottomrule
\end{tabular}
}}
\caption{Evaluation Results on ProcessBench with \textsc{Qwen2.5-Math-1.5B-NAIT}.}
\label{tab:qwen1_5_processbench}
\end{table*}

\begin{table*}[!h]
    \centering
    \small
    \resizebox{\textwidth}{!}{
    \begin{tabular}{l|cc|cccc|c}
    \toprule
    \bf Model & \bf \makecell{Training \\ Samples} & \bf \makecell{Aggregation \\ Method} & \bf GSM8K & \bf MATH & \bf \makecell{GAOKAO} & \bf \makecell{MINERVA \\ MATH} & \bf Avg. \\
    \midrule
    $\bigstar$ Qwen2.5-Math-1.5B-\textsc{MC} & 128K & Mean & 86.3 & 61.4 & 57.8 & 32.7 & 59.5 \\
    $\bigstar$ Qwen2.5-Math-1.5B-\textsc{MCRD} & 128K & Mean & 87.5 & 65.8 & 59.1 & 35.6 & 62.0 \\
    $\bigstar$ Qwen2.5-Math-1.5B-\textsc{NAIT} & 128K & Mean & 89.1 & 68.6 & 62.6 & 36.4 & 64.2 \\
    \bottomrule
    \end{tabular}
    }
    \caption{Best-of-8 evaluation performance comparison on the Reasoning benchmarks with the policy model \textsc{Qwen2.5-3B-Instruct} and \textsc{Qwen2.5-Math-1.5B}.}
    \label{tab:qwen1.5_3b}
\end{table*}

\begin{table*}[!h]
    \centering
    \small
    \resizebox{\textwidth}{!}{
    \begin{tabular}{l|cc|cccc|c}
    \toprule
    \bf Model & \bf \makecell{Training \\ Samples} & \bf \makecell{Aggregation \\ Method} & \bf GSM8K & \bf MATH & \bf \makecell{GAOKAO} & \bf \makecell{MINERVA \\ MATH} & \bf Avg. \\
    \midrule
    $\bigstar$ Qwen2.5-Math-1.5B-\textsc{MC} & 128K & Mean & 92.6 & 75.2 & 65.9 & 42.7 & 69.1 \\
    $\bigstar$ Qwen2.5-Math-1.5B-\textsc{MCRD} & 128K & Mean & 92.8 & 76.0 & 67.9 & 44.6 & 70.3 \\
    $\bigstar$ Qwen2.5-Math-1.5B-\textsc{NAIT} & 128K & Mean & 93.2 & 78.4 & 68.5 & 44.6 & 71.2 \\
    \bottomrule
    \end{tabular}
    }
    \caption{Best-of-8 evaluation performance comparison on the Reasoning benchmarks with the policy model \textsc{Qwen2.5-7B-Instruct} and \textsc{Qwen2.5-Math-1.5B}.}
    \label{tab:qwen1.5_7b}
\end{table*}

\begin{table*}[!h]
    \centering
    \small
    \resizebox{\textwidth}{!}{
    \begin{tabular}{l|cc|cccc|c}
    \toprule
    \bf Model & \bf \makecell{Training \\ Samples} & \bf \makecell{Aggregation \\ Method} & \bf GSM8K & \bf MATH & \bf \makecell{GAOKAO} & \bf \makecell{MINERVA \\ MATH} & \bf Avg. \\
    \midrule
    $\bigstar$ Qwen2.5-Math-1.5B-\textsc{MC} & 128K & Mean & 90.7 & 74.8 & 69.2 & 48.5 & 70.8 \\
    $\bigstar$ Qwen2.5-Math-1.5B-\textsc{MCRD} & 128K & Mean & 94.0 & 78.0 & 69.5 & 49.2 & 72.7 \\
    $\bigstar$ Qwen2.5-Math-1.5B-\textsc{NAIT} & 128K & Mean & 94.6 & 81.2 & 72.5 & 50.9 & 74.8 \\
    \bottomrule
    \end{tabular}
    }
    \caption{Best-of-8 evaluation performance comparison on the Reasoning benchmarks with the policy model \textsc{Qwen2.5-14B-Instruct} and \textsc{Qwen2.5-Math-1.5B}.}
    \label{tab:qwen1.5_14b}
\end{table*}

\begin{table*}[!h]
    \centering
    \small
    \resizebox{\textwidth}{!}{
    \begin{tabular}{l|cc|cccc|c}
    \toprule
    \bf Model & \bf \makecell{Training \\ Samples} & \bf \makecell{Aggregation \\ Method} & \bf GSM8K & \bf MATH & \bf \makecell{GAOKAO} & \bf \makecell{MINERVA \\ MATH} & \bf Avg. \\
    \midrule
    $\bigstar$ Qwen2.5-Math-1.5B-\textsc{MC} & 128K & Mean & 90.5 & 70.8 & 67.0 & 47.8 & 69.0 \\
    $\bigstar$ Qwen2.5-Math-1.5B-\textsc{MCRD} & 128K & Mean & 94.3 & 77.4 & 72.6 & 51.8 & 74.0 \\
    $\bigstar$ Qwen2.5-Math-1.5B-\textsc{NAIT} & 128K & Mean & 95.4 & 79.8 & 73.8 & 52.6 & 75.4 \\
    \bottomrule
    \end{tabular}
    }
    \caption{Best-of-8 evaluation performance comparison on the Reasoning benchmarks with the policy model \textsc{Qwen2.5-32B-Instruct} and \textsc{Qwen2.5-Math-1.5B}.}
    \label{tab:qwen1.5_32b}
\end{table*}

\section{PRM Evaluation Cases}
We also provide two examples of \textsc{Qwen2.5-Math-7B-NAIT} evaluations in Figure~\ref{fig:case1} and Figure~\ref{fig:case2}, demonstrating its ability to accurately distinguish between correct and incorrect reasoning steps.

\begin{figure*}[t]
\centering
\begin{tcolorbox}[
    colback=lightgray!10, 
    colframe=black, 
    title=PRM Evaluation Case 1,
    fonttitle=\bfseries
]
\begin{spacing}{1.05}
\textbf{Generator:} \textsc{Qwen2.5-14B-Instruct} \\
\textbf{Process Reward Model:} \textsc{Qwen2.5-Math-7B-NAIT} 

\vspace{0.5em}
\textbf{[Problem]} \\
A nurses’ station orders bandages in bulk packs of 50. On the first day, the nurses used 38 bandages and ordered one bulk pack of bandages. On the second day, they used ten fewer bandages. On the third day, they ordered two bulk packs of bandages and only used half a pack. They had 78 bandages left at the end of the third day. How many bandages did they start with on the first day?

\vspace{0.5em}
\textbf{[Response]} \\
To determine how many bandages the nurses started with on the first day, we need to follow these steps: \hfill \textcolor{blue}{\scriptsize [\textbf{PRM: 99.6, GT: Correct}]}

\begin{itemize}[leftmargin=1.5em, labelsep=0.5em, itemsep=0.5em]
    \item \textbf{1. Calculate the number of bandages used each day:}
    \begin{itemize}[leftmargin=1.2em, label=\(\bullet\), noitemsep]
        \item \textbf{First day:} The nurses used 38 bandages.
        \item \textbf{Second day:} They used 10 fewer bandages than the first day, so they used \(38 - 10 = 28\) bandages.
        \item \textbf{Third day:} They used half a bulk pack of bandages. Since each bulk pack contains 50 bandages, half a pack is \(50 / 2 = 25\) bandages.
    \end{itemize}
    \hfill \textcolor{blue}{\scriptsize [\textbf{PRM: 94.1, GT: Correct}]}

    \item \textbf{2. Determine the total number of bandages used over the three days:}
    \[ \text{Total bandages used} = 38 + 28 + 25 = 91 \]
    \hfill \textcolor{blue}{\scriptsize [\textbf{PRM: 98.4, GT: Correct}]}

    \item \textbf{3. Calculate the total number of bandages ordered:}
    \begin{itemize}[leftmargin=1.2em, label=\(\bullet\), noitemsep]
        \item \textbf{First day:} They ordered 1 bulk pack, which is 50 bandages.
        \item \textbf{Third day:} They ordered 2 bulk packs, which is \(2 \times 50 = 100\) bandages.
        \item \textbf{Total bandages ordered:} \(50 + 100 = 150\).
    \end{itemize}
    \hfill \textcolor{blue}{\scriptsize [\textbf{PRM: 83.2, GT: Correct}]}

    \item \textbf{4. Determine the total number of bandages available at the beginning:}
    \begin{itemize}[leftmargin=1.2em, label=\(\bullet\), noitemsep]
        \item At the end of the third day, they had 78 bandages left.
        \item Let \(B\) be the initial number of bandages.
        \item The total number of bandages used plus the remaining bandages equals the initial number of bandages:
             \[ B = \text{Total bandages used} + \text{Remaining bandages} \]
             \[ B = 91 + 78 = 169   \]
    \end{itemize}
    \hfill \textcolor{red}{\scriptsize [\textbf{PRM: 0.5, GT: Incorrect}]}
\end{itemize}

Thus, the nurses started with \fbox{169} bandages on the first day. \hfill \textcolor{red}{\scriptsize [\textbf{PRM: 29.4, GT: Incorrect}]}

\end{spacing}
\end{tcolorbox}
\caption{PRM Evaluation Case 1}
\label{fig:case1}
\end{figure*}

\begin{figure*}[t]
\centering
\begin{tcolorbox}[
    colback=lightgray!10, 
    colframe=black, 
    title=PRM Evaluation Case 2,
    fonttitle=\bfseries
]
\begin{spacing}{1.05}
\textbf{Generator:} \textsc{Qwen2.5-14B-Instruct} \\
\textbf{Process Reward Model:} \textsc{Qwen2.5-Math-7B-NAIT} 

\vspace{0.5em}
\textbf{[Problem]} \\
Kim sleepwalks, to monitor her sleeping hours, she installs a camera in her room. Kim usually goes to sleep at 10 p.m. and wakes up at 6 a.m. She is a sleepwalker. One day, after reviewing the cameras, she finds that she woke up that day and was sleepwalking from 2:15 to 2:38 am.  Also, that day she woke up 5 minutes earlier than usual to go to the bathroom. How many minutes did she sleep on her bed that day?

\vspace{0.5em}
\textbf{[Response]} \\
To determine how many minutes Kim slept on her bed that day, we need to account for her usual sleep schedule, the time she spent sleepwalking, and the additional time she woke up early. \hfill \textcolor{blue}{\scriptsize [\textbf{PRM: 99.6, GT: Correct}]}

\begin{itemize}[leftmargin=1.5em, labelsep=0.5em, itemsep=0.5em]
    \item \textbf{1. Calculate the total sleep time if there were no interruptions:}
    \begin{itemize}[leftmargin=1.2em, label=\(\bullet\), noitemsep]
        \item Kim normally sleeps from 10 p.m. to 6 a.m., which is:
            \[     8 \text{ hours} = 480 \text{ minutes}     \]
    \end{itemize}
    \hfill \textcolor{blue}{\scriptsize [\textbf{PRM: 99.6, GT: Correct}]}

    \item \textbf{2. Account for the time she woke up early:}
    \begin{itemize}[leftmargin=1.2em, label=\(\bullet\), noitemsep]
        \item Kim woke up 5 minutes earlier than usual. This means she was awake for an additional 5 minutes before her usual wake-up time of 6 a.m. Therefore, she lost 5 minutes of sleep:
            \[480 \text{ minutes} - 5 \text{ minutes} = 475 \text{ minutes}\]
    \end{itemize}
    \hfill \textcolor{blue}{\scriptsize [\textbf{PRM: 88.3, GT: Correct}]}

    \item \textbf{3. Subtract the time spent sleepwalking:}
    \begin{itemize}[leftmargin=1.2em, label=\(\bullet\), noitemsep]
        \item Kim was sleepwalking from 2:15 a.m. to 2:38 a.m., which is:
        \[     2:38 \text{ a.m.} - 2:15 \text{ a.m.} = 23 \text{ minutes}     \]
        \item Since sleepwalking does not count as sleep on her bed, we subtract these 23 minutes from the remaining sleep time:
        \[     475 \text{ minutes} - 23 \text{ minutes} = 452 \text{ minutes}     \]
    \end{itemize}
    \hfill \textcolor{blue}{\scriptsize [\textbf{PRM: 92.6, GT: Correct}]}
\end{itemize}
Therefore, the number of minutes Kim slept on her bed that day is \fbox{452}. \hfill \textcolor{blue}{\scriptsize [\textbf{PRM: 99.3, GT: Correct}]}

\end{spacing}
\end{tcolorbox}
\caption{PRM Evaluation Case 2}
\label{fig:case2}
\end{figure*}

\label{sec:appendix}

\end{document}